\documentclass[letterpaper]{article} % DO NOT CHANGE THIS
\usepackage{aaai23}  % DO NOT CHANGE THIS
\usepackage{times}  % DO NOT CHANGE THIS
\usepackage{helvet}  % DO NOT CHANGE THIS
\usepackage{courier}  % DO NOT CHANGE THIS
\usepackage[hyphens]{url}  % DO NOT CHANGE THIS
\usepackage{graphicx} % DO NOT CHANGE THIS
\usepackage{natbib}  % DO NOT CHANGE THIS AND DO NOT ADD ANY OPTIONS TO IT
\usepackage{caption} % DO NOT CHANGE THIS AND DO NOT ADD ANY OPTIONS TO IT
\usepackage{algorithm}
\usepackage{algorithmicx}
\usepackage{comment}
\usepackage{booktabs} % For formal tables
\usepackage{amsmath,bm,amssymb}
\usepackage{algpseudocode}
\usepackage{longtable}
\usepackage{subfigure}
\usepackage{cases}
\usepackage{diagbox}
\usepackage{mathrsfs}
\usepackage{multirow}
\usepackage{caption}
\usepackage{color}
\usepackage{cite}
\usepackage{makecell}
\usepackage{wrapfig}
\usepackage{newfloat}
\usepackage{listings}
\DeclareCaptionStyle{ruled}{labelfont=normalfont,labelsep=colon,strut=off} % DO NOT CHANGE THIS
\floatstyle{ruled}
\newfloat{listing}{tb}{lst}{}
\floatname{listing}{Listing}
\title{Adversarial Alignment for Source Free Object Detection}
\author{
    Qiaosong Chu\textsuperscript{\rm 1,2}\equalcontrib,
    Shuyan Li\textsuperscript{\rm 1,2}\equalcontrib,
    Guangyi Chen\textsuperscript{\rm 3,4},
    Kai Li\textsuperscript{\rm 5},
    Xiu Li\textsuperscript{\rm 1,2}\thanks{*Corresponding author} 
}
\affiliations{
    \textsuperscript{\rm 1}Tsinghua Shenzhen International Graduate School, Shenzhen, China\\
    \textsuperscript{\rm 2}Tsinghua University, Beijing, China\\
    \textsuperscript{\rm 3}Carnegie Mellon University, Pittsburgh PA, USA\\
    \textsuperscript{\rm 4}Mohamed bin Zayed University of Artificial Intelligence, Abu Dhabi, UAE\\
    \textsuperscript{\rm 5}NEC LABORATORIES AMERICA, INC\\
    \ zqs20@mails.tsinghua.edu.cn,
    sl2141@cam.ac.uk,
    \{guangyichen1994, li.gml.kai\}@gmail.com,
    li.xiu@sz.tsinghua.edu.cn
}

\usepackage{bibentry}
\begin{document}

\maketitle

\begin{abstract}
Source-free object detection~(SFOD) aims to transfer a detector pre-trained on a label-rich source domain to an unlabeled target domain without seeing source data. While most existing SFOD methods generate pseudo labels via a source-pretrained model to guide training, these pseudo labels usually contain high noises due to heavy domain discrepancy. In order to obtain better pseudo supervisions, we divide the target domain into source-similar and source-dissimilar parts and align them in the feature space by adversarial learning.
Specifically, we design a detection variance-based criterion to divide the target domain. This criterion is motivated by a finding that larger detection variances denote higher recall and larger similarity to the source domain. Then we incorporate an adversarial module into a mean teacher framework to drive the feature spaces of these two subsets indistinguishable. Extensive experiments on multiple cross-domain object detection datasets demonstrate that our proposed method consistently outperforms the compared SFOD methods. 
Our implementation is available at https://github.com/ChuQiaosong.  
\end{abstract}

\section{Introduction}

Despite the promising performance, deep object detection still heavily relies on numerous manually annotated training data. It leads to a significant performance drop in real-world scenarios when the detection system faces a new environment with the domain shift. As collecting labels for all conditions is impractical, it requires detectors to efficiently adapt to new environments without further annotations. To this end, Unsupervised Domain Adaptive~(UDA) object detection has gained increasing attention in recent years~\cite{DBLP:conf/iccv/HeZ19,DBLP:conf/cvpr/SaitoUHS19,DBLP:conf/iccv/WuLHZ021}.

\begin{figure}
\centering
\includegraphics[width=0.9\linewidth]{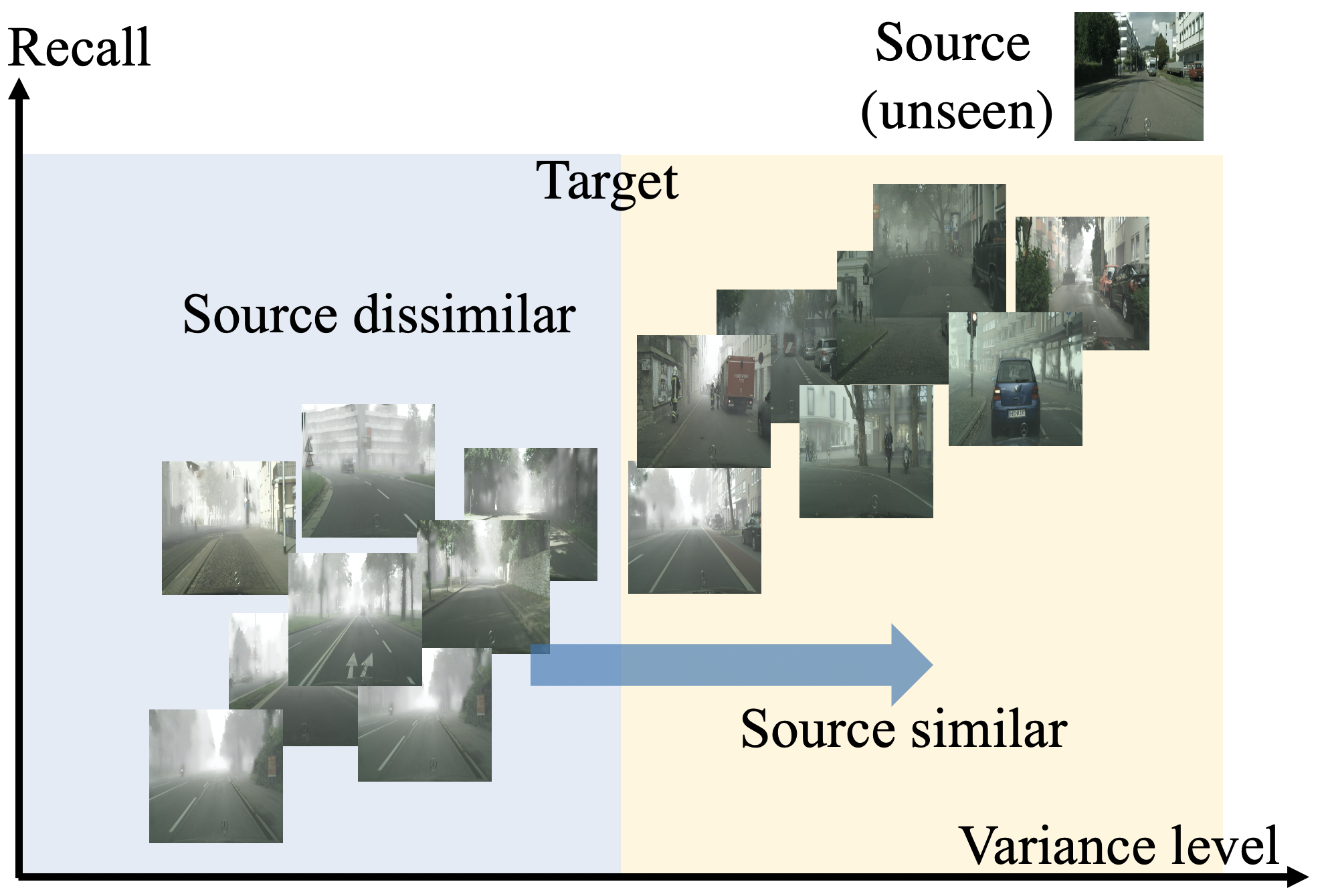}
\caption{Basic idea of $\text{A}^2$SFOD. We propose a variance-based criterion to divide the target domain data into source-similar and source-dissimilar subsets, based on a finding that larger detection variances denote higher recall and larger similarity to the source domain. We can achieve the domain alignment by simulating the source and target domains with the divided source-similar and source-dissimilar subsets.}
\label{Fig:idea}
\end{figure}

The aforementioned methods are based on the assumption that both source and target domain data are accessible. However, this assumption may not hold in many real-world applications due to infeasible data transmission, computation resource restrictions, or data privacy. It poses a new challenge, which is named source data-free or source-free, i.e. only a well-trained source model is provided during adapting to the target domain without having access to source data~\cite{DBLP:conf/icml/KunduKBMKJR22,Wang_2022_CVPR,Yazdanpanah_2022_CVPR, Machireddy_2022_CVPR}.

While the source-free challenge has been well studied for image classification tasks~\cite{DBLP:conf/iccv/XiaZD21, DBLP:conf/icml/LiangHF20}, there are much fewer works that focus on Source-Free Object Detection~(SFOD)~\cite{DBLP:conf/mmasia/ZhangYXLL21,DBLP:journals/corr/abs-2203-15793}. Due to complex background, obscured objects, and numerous negative samples, directly applying conventional source-free domain adaptation methods to SFOD cannot achieve satisfactory detection accuracy. Therefore, it is desirable to develop effective domain adaptation methods to solve the source-free problem for object detection. 

As there is no manually labeled data available during adaptation, most existing SFOD methods train the model by using pseudo-labels generated by a source-pretrained model~\cite{DBLP:conf/icassp/YuanCYXXZP22, DBLP:conf/mmasia/ZhangYXLL21}. However, the domain shift inevitably introduces high noises in pseudo labels, which deteriorates the detection performance~\cite{DBLP:conf/cvpr/Deng0CD21}.
Though various data augmentation methods~\cite{DBLP:conf/aaai/LiCXYYPZ21} have been developed to improve the quality of pseudo labels, the domain discrepancy has not been well narrowed. The source-pretrained knowledge is difficult to adapt to these hard samples far dissimilar from the source data.

To address this problem, we propose an adversarial learning based source free object detection method~($\text{A}^2$SFOD). As shown in Figure~\ref{Fig:idea}, we aim to drive the source-dissimilar features close to the source-similar ones, such that pseudo labels generated by the source-pretrained model are of high quality over the whole target domain. To this end, we design a detection variance-based criterion to separate the target domain data into source-similar ones and source-dissimilar ones. This criterion is motivated by a finding that larger detection variances denote higher recall and larger similarity to the source domain. Given source-similar and source-dissimilar subsets, we propose to apply adversarial learning to the mean teacher structure to learn a model for feature space alignment. We conduct extensive experiments on five widely used detection datasets to validate the superiority of our method. 
% Initializing both teacher and student models with the aligned parameters, we further finetune the mean teacher network with all target data with mosaic augmentation to improve the detection of small and obscured objects.

The contribution of this paper can be summarized as: 1) we find the detection variance and the similarity to source data are positively correlated, and further propose a criterion to divide the target domain; 2) we present an adversarial alignment method to refine the feature space to obtain pseudo labels of higher quality; 3) we achieve consistent and significant improvement on 5 datasets and 4 settings.

\section{Related Work}
Domain Adaptive Object Detection (DAOD)~\cite{DBLP:conf/cvpr/CaiPNTDY19,DBLP:conf/cvpr/Chen0SDG18,DBLP:conf/cvpr/XuWNTZ20,DBLP:conf/cvpr/GuSX20,DBLP:conf/icig/ZhangMW21,DBLP:journals/corr/abs-2108-00977} aims to address the domain shift problems in object detection task. Existing DAOD methods can be divided into two categories: feature alignment methods and self-training methods. The former ones aim to align source domain and target domain by learning a domain-agnostic feature space~\cite{DBLP:conf/cvpr/Chen0SDG18,DBLP:conf/cvpr/SaitoUHS19,DBLP:conf/cvpr/ChenZD0D20,DBLP:journals/corr/abs-2111-13216,DBLP:conf/cvpr/Zheng0LW20,DBLP:journals/tip/WangZZCT22}. For example DA-Faster~\cite{DBLP:conf/cvpr/Chen0SDG18} first proposed to tackle domain shift on image-level and instance-level to learn a domain-invariant region proposal network (RPN). SWDA~\cite{DBLP:conf/cvpr/SaitoUHS19} attempted to align distributions of foreground objects rather than the whole image based on strong local alignment and weak global alignment. The latter ones train the model recursively by using self-training to gradually increase the accuracy of generated pseudo labels on the target domain~\cite{DBLP:conf/cvpr/InoueFYA18, DBLP:conf/iccv/KhodabandehVRM19,DBLP:conf/iccv/KimCKK19,DBLP:conf/cvpr/RoyChowdhuryCSJ19}. These methods vary by different strategies to refine pseudo labels and update models. For example, WST~\cite{DBLP:conf/iccv/KimCKK19} proposed weak self-training for stable training and designed adversarial background score regularization to tackle domain shifts. NL~\cite{DBLP:conf/iccv/KhodabandehVRM19} formulated domain adaption as training with noisy labels and refined pseudo labels by using a classification module.

In real-world scenarios, the source data is usually inaccessible due to data privacy, leading to the SFOD problem~\cite{DBLP:conf/icml/0004JYY22,DBLP:journals/corr/abs-2207-05785,DBLP:journals/corr/abs-2205-14566, DBLP:journals/corr/abs-2205-12840, DBLP:journals/corr/abs-2205-10711}. 
Due to complex background and negative examples, SFOD is far more challenging than conventional source-free image classification~\cite{DBLP:conf/wacv/AgarwalPZG22, DBLP:journals/corr/abs-2208-04226,  DBLP:journals/corr/abs-2207-07624,DBLP:conf/iccv/XiaZD21}.
SFOD-Mosaic~\cite{DBLP:conf/aaai/LiCXYYPZ21} first formulated the SFOD problem and proposed to search for a fairly good confidence threshold and enabled self-training via generated pseudo labels. SOAP~\cite{DBLP:journals/ijis/XiongYZGLZ21} then added a domain-specific perturbation on the target domain and optimized the source-pretrained model via self‐supervised learning. HCL~\cite{DBLP:conf/nips/HuangGXL21} exploited historical source hypothesis to make up for the lack of source data. S\&M~\cite{DBLP:conf/icassp/YuanCYXXZP22} proposed a Simulation-and-Mining framework that modeled the SFOD task into an unsupervised false negatives mining problem. Recently, more new methods have been developed~\cite{DBLP:conf/cvpr/Li0ZZX22,DBLP:journals/pami/LiangHWHF22}. However, these methods cannot well narrow the gap between the source domain and the target domain. All in all, SFOD is far from being fully explored and more effective SFOD methods are desired to develop.  
\begin{figure*}[t]
\centering
\includegraphics[width=0.9\linewidth]{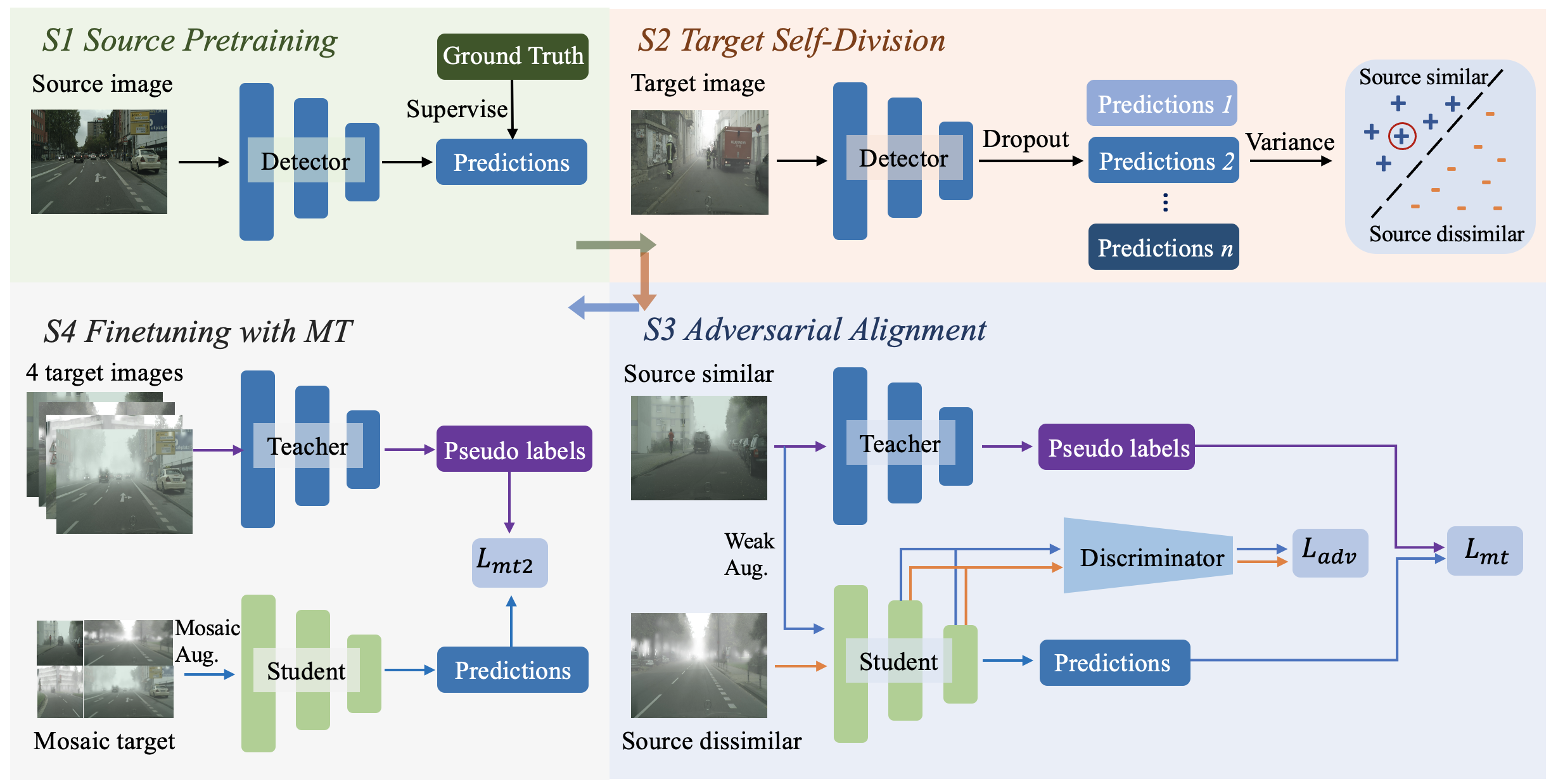}
\caption{Framework of $\text{A}^2$SFOD. It contains four stages, source pre-training, target self-division, adversarial alignment, and fine-tuning with MT, where the first stage is unseen during adaptation.}
\label{Fig:framework}
% \vspace{-0.3cm} 
\end{figure*}

\section{Method}
Source-free object detection (SFOD) aims to adapt a detector pre-trained on the source domain to an unlabeled target domain. In this process, the data in the source domain is untouched.  
Specifically, given an unlabeled target dataset $\{X^t_i\}^N_{i=1}$ ($N$ is the number of images) and a detector $F$ with source pre-trained parameters $\theta_s$~(e.g. a Faster-RCNN~\cite{DBLP:conf/nips/RenHGS15} model), we need to update the parameters to $\theta_t$ for the target domain.
In this paper, we propose a method for this task, called $\text{A}^2$SFOD, whose overall framework is shown in Figure~\ref{Fig:framework}.
In $\text{A}^2$SFOD, we first divide the target data into two subsets, source-similar and source-dissimilar, via the variance of the predictions from the source-pretrained detector $F_{\theta_s}$. Then we align these two subsets via adversarial learning and finetune the detector via mean-teacher learning. In the following sections, we will detail each stage.

\subsection{Target Self-Division}

Aligning source and target domain is a widely-used method for conventional domain adaptation tasks~\cite{DBLP:conf/cvpr/Kang0YH19,DBLP:conf/cvpr/ZhuPYSL19,DBLP:conf/cvpr/0007CGV19}. Given the source and target data, we can intuitively achieve the alignment in the data space~\cite{DBLP:conf/cvpr/ChenZD0D20} or feature space~\cite{DBLP:conf/cvpr/SaitoUHS19}. However, the lack of source data poses a new challenge for domain alignment.

Although we have no access to source data, the source-pretrained model does convey rich information about the source domain. Accordingly, we propose to self-divide the target data into two subsets by the pre-trained model to explicitly simulate the source and target domain. To achieve this goal, we design a detection variance-based division criterion, where the detection variance is calculated based on predictions yielded by the source-pretrained model on target data. It is motivated by a finding that larger detection variances denote higher similarity to the source data. Specifically, the pre-trained model tends to yield more predictions of hard samples for source-similar images while directly ignore these hard samples for source-dissimilar images. These predictions of hard samples usually have high uncertainty, and hence have larger variances during adaptation. 

We formulate the calculation of the detection variance as:
\begin{equation}
\begin{aligned}
v_i &= E[(F_{\theta_s}(X_i) - E[F_{\theta_s}(X_i)])^2],
% &= \int_{\theta}[F(X_i) - \int_{\theta_0}F(X_i|\theta_0)p(\theta_0)d\theta_0]^2p(\theta)d\theta,
\label{Equ:vi}
\end{aligned}
\end{equation}
where $F_{\theta_s}(X_i)$ represents the predictions of image $X_i$ via the source-pretrained model.
% and $p(\theta)$ is the probability density function of $\theta$. 
As such calculation is intractable in practice, we instead approximate this calculation with Monte-Carlo sampling. Inspired by \cite{DBLP:conf/icml/GalG16}, we formulate the sampling function with dropout, which is a widely-used stochastic regularization tool in deep learning~\cite{DBLP:journals/corr/BlundellCKW15}. This approximation is easy to perform via $M$ stochastic forward passes without changing the detection model. 

As the corresponding outputs $F_{\theta_s}(X_i)=(\boldsymbol{b_i}, \boldsymbol{c_i})$ are composed of localization coordinates and classification scores, we formulate the detection variance as the product of two terms, box localization variance $v_{bi}$ and classification score variance $v_{ci}$. Given a prediction with $N_i$ boxes and $K$ classes, we have $\{\boldsymbol{b_{ij}} = (x_{ij}^1, y_{ij}^1, x_{ij}^2, y_{ij}^2)\}^{N_i}_{j=1}$ and $\{\boldsymbol{c_{ij}} = (c_{ij}^1, c_{ij}^2, ..., c_{ij}^K)\}^{N_i}_{j=1}$. We can formulate $v_{bi}$ and $v_{ci}$ as follows:

\begin{equation}
v_{bi} = \frac{1}{MN_i}\sum\limits^{N_i}_{j=1}\sum\limits^{M}_{m=1}||\boldsymbol{b_{ij}^m} - \overline{\boldsymbol{b_{ij}}}||^2,
\end{equation}
\begin{equation}
v_{ci} = \frac{1}{MN_i}\sum\limits^{N_i}_{j=1}\sum\limits^{M}_{m=1}||\boldsymbol{c_{ij}^m} - \overline{\boldsymbol{c_{ij}}}||^2,
\end{equation}
where $\boldsymbol{b_{ij}^m}$, $\boldsymbol{c_{ij}^m}$ denote the localization coordinates and classification scores of the $m$-th forward pass of the $j$-th bounding box in $X_i$ respectively, and $\overline{\boldsymbol{b_{ij}}}$, $\overline{\boldsymbol{c_{ij}}}$ denote the corresponding average value of total $M$ forward passes. Then we have the detection variance of $X_i$ as $v_i = v_{bi} v_{ci}$. We order these images according to their variances from small to large, and use $r_i$ to denote the ranking of $X_i$. We define the variance level of the $i$-th image as $vl_i=\frac{r_i}{N}$. We consider $X_i$ as source-similar if $vl_i\geq\sigma$ and source-dissimilar otherwise, where $\sigma\in (0,1)$ is a pre-defined threshold. In this way, we divide the target domain data into source-similar and source-dissimilar subsets for the preparation of adversarial alignment.  

\subsection{Adversarial Alignment}
In this subsection, we introduce how to achieve domain alignment with the divided source similar and source dissimilar subsets. As shown in Stage 3 in Figure~\ref{Fig:framework}, we incorporate an adversarial training procedure into a mean-teacher architecture to achieve domain alignment in the feature space.
Specifically, we build a teacher model $F_{tea}$ and a student model $F_{stu}$, which apply the same network architecture with pretrained model $F$ and are initialized with parameters $\theta_s$. 
The parameters of the student model are quickly updated for domain alignment while the parameters of the teacher model are slowly updated.
There are two loss functions in our adversarial alignment process to learn the student model: mean teacher loss for model adaptation and adversarial loss for domain alignment.

The goal of the mean teacher loss is to use the pseudo labels generated with the pretrained teacher model to supervise the training of the student model. First, we feed the teacher model with source similar data $X_i^s$ and feed the student model with the data augmentation version $Aug(X_{i}^s)$. Then the mean teacher loss is formulated with the outputs of both teacher and student models as:
\begin{equation}
\begin{aligned}
L_{mt} &= L_{cls}(F_{stu}(Aug(X^s_{i}), F_{tea}(X^s_i))) \\
& + L_{reg}(F_{stu}(Aug(X^s_{i}), F_{tea}(X^s_i))),
\end{aligned}
\label{Equ:mt}
\end{equation}
where $L_{reg}$ is the location regression loss which calculates the L1-smooth distance for predicted and supervised bounding box
and $L_{cls}$ is the cross-entropy loss for classification supervision. Both $L_{reg}$ and $L_{cls}$ are widely-used losses in the field of object detection, such as Faster-RCNN~\cite{DBLP:conf/nips/RenHGS15}.

To align the feature spaces of source similar and source dissimilar subsets, we further conduct adversarial learning with the outputs of the student model. We take the student model as the “generator” and build extra ``discriminators'' to play a min-max adversarial game, where the ``generator'' tries to fool the ``discriminators'' by generating features that can't be distinguished as source similar or source dissimilar. To capture both local and global information, we follow SW-Faster~\cite{DBLP:conf/cvpr/SaitoUHS19} to build a local discriminator $D_l$ and a global discriminator $D_g$, where $D_l$ focus on the foreground objects and $D_g$ focus on the background.

The adversarial losses with global and local discriminators can be formulated as:
% We designed the global adversarial loss and local adversarial loss as:
\begin{equation}
\begin{aligned}
L_{local} &=\frac{1}{WHN_s}\sum\limits^{N_s}_{i=1}\sum\limits^{W}_{w=1}\sum\limits^{H}_{h=1}D_l(F_l(X_i^s))^2_{wh} \\
&+ \frac{1}{WHN_d}\sum\limits^{N_d}_{j=1}\sum\limits^{W}_{w=1}\sum\limits^{H}_{h=1}(1 - D_l(F_l(X_j^d))^2_{wh},
\end{aligned}
\end{equation}
\begin{equation}
\begin{aligned}
L_{global} =& -\frac{1}{N_s}\sum\limits^{N_s}_{i=1}(1 - D_g(F_g(X_i^s)))^{\gamma}log(D_g(F_g(X_i^s)))  \\ 
& -\frac{1}{N_d}\sum\limits^{N_d}_{j=1}D_g(F_g(X_j^d))^{\gamma}log(1 - D_g(F_g(X_j^d))),
\end{aligned}
\label{Equ:labda}
\end{equation}
where $X_i^s$, $X_j^d$ denote the $i$-th source similar image and $j$-th source dissimilar image; $F_l$, $F_g$ denote different layers on the backbone that capture the local feature (feature maps) and global feature (a feature vector) respectively; $W, H$ denote width and height of local feature map on the backbone; $N_s$, $N_d$ denote the total number of source similar images and source dissimilar images, and $\gamma$ is a Focal loss~\cite{lin2017focal} parameter which controls the model to focus on hard-to-classify examples but not the easy ones. 
Compared with the local adversarial loss, the global one applies the Focal loss to focus on the hard examples and ignore easy-to-classify examples to achieve a weak alignment, without hurting the performance of the local model~\cite{DBLP:conf/cvpr/SaitoUHS19}.

Finally, we update the student model by fusing the 
mean teacher loss and adversarial losses as
\begin{equation}
\max\limits_{D}\min\limits_{F}L_{mt} - \lambda L_{adv},
\label{Equ:all}
\end{equation}
where $L_{adv} = L_{local} + L_{global}$. With this loss function, these embeddings are encouraged to have similar distributions for both source-similar and source-dissimilar images (achieved by adversarial loss $L_{adv}$) and have a great discriminative ability for the target domain detection (achieved by mean teacher loss $L_{mt}$).

\subsection{Fine-tuning with Mean Teacher}
After adversarial alignment, we can get pseudo labels of high quality over the whole target data. Hence, we finetune the detector by using both source-similar and source-dissimilar data to make full use of the information of the whole target domain. Both teacher and student models are initialized with the parameters of the student model learned in stage 3. 
As the detection error mainly comes from false negative objects~\cite{DBLP:conf/aaai/LiCXYYPZ21}, we employ mosaic augmentation~\cite{DBLP:journals/corr/abs-2203-15793,DBLP:conf/bmvc/WangL0K021} to simulate the false negatives to better detect small-scale and obscured objects. 

As shown in Figure~\ref{Fig:framework} Stage 4, we feed the teacher model with four independent target images $\{X_{i1}^t,X_{i2}^t,X_{i3}^t,X_{i4}^t\}$ and generate independent predictions $\{Y_{i1}^t,Y_{i2}^t,Y_{i3}^t,Y_{i4}^t\}$. Then we resize and mosaic these four images with data augmentation into a combined image $X_{im}$, with the same size as the original input $X_{i1}^t$. Likewise, we obtain the mosaic pseudo label $Y_{im}$. We formulate the loss during the fine-tuning stage as follows:
\begin{equation}
L_{mt2} = L_{cls}(F_{stu}(X_{im}), Y_{im}) + L_{reg}(F_{stu}(X_{im}), Y_{im}),
\end{equation}
where $L_{reg}$ and $L_{cls}$ are basic location regression and classification losses respectively, which are the same as the ones in Equ.~(\ref{Equ:mt}).

\par
\section{Experiments}

We have conducted extensive experiments to evaluate our method, including comparisons with other SFOD methods, detailed ablation studies, and analysis.

\subsection{Datasets}
We evaluated our method on five popular object detection datasets. The detailed information of these datasets is summarized in the following: \textbf{(1)Cityscapes}~\cite{DBLP:conf/cvpr/CordtsORREBFRS16} collects different scenes from various cities on the street, which contains 2,975 training images and 500 validation images. We utilized the rectangle of the instance mask to obtain bounding boxes following previous work. \textbf{(2)Foggy-Cityscapes}~\cite{DBLP:conf/eccv/SakaridisDHG18} is constructed from Cityscapes by simulating three levels of foggy weather. It contains the same amount of images as the Cityscapes. We inherited the annotations of Cityscapes. \textbf{(3)KITTI}~\cite{DBLP:conf/cvpr/GeigerLU12} is a dataset containing 7,481 training images for autonomous driving different from Cityscapes. \textbf{(4)Sim10k}~\cite{DBLP:conf/icra/Johnson-Roberson17} is a simulation dataset generated from a popular computer game Grand Theft Auto V. It contains 10,000 images of the synthetic driving scene with 58,071 bounding boxes of the car. \textbf{(5)BDD100k}~\cite{DBLP:journals/corr/abs-1805-04687} is a large dataset including 100k images with six types of weather, six different scenes, and three categories for the time of day. Following~\cite{DBLP:conf/cvpr/XuZJW20}, we applied the daytime subset of the dataset in our experiment, where 36,728 images were used for training and the other 5,258 images for validation.

\begin{table*}[t]
\begin{center}
% \vspace{-0.1cm}
% \small
\centering
\begin{tabular}{c|cccccccc|c}
\Xhline{0.4pt}
Methods & truck & car & rider & person & train & motor & bicycle & bus & mAP \\
\Xhline{0.4pt}
Source only & 10.7 & 30.2 & 30.8 & 23.6 & 9.2 & 16.3 & 24.7 & 19.7 & 20.6 \\
\Xhline{0.4pt}
DA-Faster~\cite{DBLP:conf/cvpr/Chen0SDG18} & 19.5 & 43.5 & 36.5 & 28.7 & 12.6 & 24.8 & 29.1 & 33.1 & 28.5 \\
SW-Faster~\cite{DBLP:conf/cvpr/SaitoUHS19} & 23.7 & 47.3 & 42.2 & 32.3 & 27.8 & 28.3 & 35.4 & 41.3 & 34.8 \\
MAF~\cite{DBLP:conf/iccv/HeZ19} & 23.8 & 43.9 & 39.5 & 28.2 & 33.3 & 29.2 & 33.9 & 39.9 & 34.0 \\
CR-DA-DET~\cite{DBLP:conf/cvpr/XuZJW20} & 27.2 & 49.2 & 43.8 & 32.9 & 36.4 & 30.3 & 34.6 & 45.1 & 37.4 \\
AT-Faster~\cite{DBLP:conf/eccv/HeZ20} & 23.7 & 50.0 & 47.0 & 34.6 & 38.7 & 33.4 & 38.8 & 43.3 & 38.7 \\
HCL~\cite{DBLP:conf/nips/HuangGXL21} & 26.9 & 46.0 & 41.3 & 33.0 & 25.0 & 28.1 & 35.9 & 40.7 & 34.6 \\
\Xhline{0.4pt}
SFOD-Mosaic~\cite{DBLP:conf/aaai/LiCXYYPZ21} & 25.5 & 44.5 & 40.7 & \textbf{33.2} & 22.2 & 28.4 & 34.1 & \textbf{39.0} & 33.5 \\
$\text{A}^2$SFOD(Ours) & \textbf{28.1} & \textbf{44.6} & \textbf{44.1} & 32.3 & \textbf{29.0} & \textbf{31.8} & \textbf{38.9} & 34.3 & \textbf{35.4} \\
\Xhline{0.4pt}
Oracle & 38.1 & 49.8 & 53.1 & 33.1 & 37.4 & 41.1 & 57.4 & 48.2 & 44.8 \\
\Xhline{0.4pt}
\end{tabular}
\end{center}
\caption{Adaptation from Normal to Foggy Weather: Cityscapes $\rightarrow$ Foggy-Cityscapes}
\label{Tab:Cityscapes-Foggy-Cityscapes}
% \vspace{-0.3cm}
\end{table*}

\subsection{Implementation Details}
We followed the setting in~\cite{DBLP:conf/cvpr/Chen0SDG18} that adopted Faster-RCNN~\cite{DBLP:conf/nips/RenHGS15} with VGG-16 pretrained on ImageNet~\cite{DBLP:journals/ijcv/RussakovskyDSKS15} for our detection model. In all experiments, the shorter side of each input image was resized to 600. The detector was trained with Stochastic Gradient Descent (SGD) with a learning rate of 0.001. To stabilize the training of Mean Teacher~\cite{DBLP:conf/nips/TarvainenV17},
we only updated the teacher model every 2500 iterations using exponential moving average (EMA) weights of the student model. In the pseudo label generation process, we filtered out the bounding boxes whose classification scores were lower than 0.7 to control the quality of pseudo labels.
In the testing phase, we evaluated the adaptation performance by reporting mean average precision (mAP) with an IoU threshold of 0.5. Following~\cite{DBLP:conf/cvpr/SaitoUHS19}, we set $\lambda$ = 0.1 for Sim10k $\rightarrow$ Cityscapes in Equ.~(\ref{Equ:all}) and $\lambda$ = 1 for other tasks. We set the threshold parameter $\sigma=0.7$ as it is empirically found to result in the best performance. All experiments were implemented with PyTorch 1.7.1. 

\begin{table}[tbp]
% \vspace{-0.1cm}
\begin{center}
\centering
\begin{tabular}{c|c}
\Xhline{0.4pt}
Methods & AP of car \\
\Xhline{0.4pt}
Source only & 35.7 \\
\Xhline{0.4pt}
DA-Faster~\cite{DBLP:conf/cvpr/Chen0SDG18} & 38.5 \\
SW-Faster~\cite{DBLP:conf/cvpr/SaitoUHS19} & 37.9 \\
MAF~\cite{DBLP:conf/iccv/HeZ19} & 41.0 \\
AT-Faster~\cite{DBLP:conf/eccv/HeZ20} & 42.1 \\
Noise Labeling~\cite{DBLP:conf/iccv/KhodabandehVRM19} & 43.0 \\
SOAP~\cite{DBLP:journals/ijis/XiongYZGLZ21} & 42.7 \\
\Xhline{0.4pt}
SFOD-Mosaic~\cite{DBLP:conf/aaai/LiCXYYPZ21} & 44.6 \\
$\text{A}^2$SFOD(Ours) & \textbf{44.9} \\
\Xhline{0.4pt}
Oracle & 58.5 \\
\Xhline{0.4pt}
\end{tabular}
\end{center}
% \vspace{-0.3cm}
\caption{Adaptation to a New Sense: KITTI $\rightarrow$ Cityscapes}
\label{Tab:KITTI-Cityscapes}
\end{table}

\subsection{Comparisons with Other SFOD methods}

In this subsection, we evaluated the transferability of our method in 4 aspects, including from a normal environment to foggy weather, from training dataset to unseen new scenes, from synthetic to real images, and from a data-limited source to a large-scale target. 
For a fair comparison, we strictly followed the experiment setting of SFOD-Mosaic~\cite{DBLP:conf/aaai/LiCXYYPZ21} and applied the similar source-only model, even if a more complex source-only model results in better performance. Specifically, we mainly compared our method $\text{A}^2$SFOD with multiple recent methods such as DA-Faster~\cite{DBLP:conf/cvpr/Chen0SDG18}, SW-Faster~\cite{DBLP:conf/cvpr/SaitoUHS19}, DA-Detection~\cite{DBLP:conf/wacv/HsuYTHT0020}, MAF~\cite{DBLP:conf/iccv/HeZ19}, AT-Faster~\cite{DBLP:conf/eccv/HeZ20}, and Noise Labeling~\cite{DBLP:conf/iccv/KhodabandehVRM19}; the baseline method SFOD-Mosaic~\cite{DBLP:conf/aaai/LiCXYYPZ21}; the ``Source only'' method trained with only source training data as the lower bound; and the ``Oracle'' method trained using labeled target data as the upper bound. Generally speaking, we achieved significant performance improvement over other methods in all settings.

\begin{table}[tbp]
% \vspace{-0.1cm}
\begin{center}
\centering
\begin{tabular}{c|c}
\Xhline{0.4pt}
Methods & AP of car \\
\Xhline{0.4pt}
Source only & 33.7 \\
\Xhline{0.4pt}
DA-Faster~\cite{DBLP:conf/cvpr/Chen0SDG18} & 38.5 \\
SW-Faster~\cite{DBLP:conf/cvpr/SaitoUHS19} & 40.1 \\
MAF~\cite{DBLP:conf/iccv/HeZ19} & 41.1 \\
AT-Faster~\cite{DBLP:conf/eccv/HeZ20} & 42.1 \\
HTCN~\cite{DBLP:conf/cvpr/ChenZD0D20} & 42.5 \\
Noise Labeling~\cite{DBLP:conf/iccv/KhodabandehVRM19} & 43.0 \\
\Xhline{0.4pt}
SFOD-Mosaic~\cite{DBLP:conf/aaai/LiCXYYPZ21} & 43.1 \\
$\text{A}^2$SFOD(Ours) & \textbf{44.0} \\
\Xhline{0.4pt}
Oracle & 58.5 \\
\Xhline{0.4pt}
\end{tabular}
\end{center}
% \vspace{-0.3cm}
\caption{Adaptation from Synthetic to Real Images: Sim10k $\rightarrow$ Cityscapes}
\label{Tab:Sim10k-Cityscapes}
\end{table}

\begin{table*}[t]
% \vspace{-0.1cm}
\begin{center}
% \small
\centering
\begin{tabular}{c|ccccccc|c}
\Xhline{0.4pt}
Methods & truck & car & rider & person & motor & bicycle & bus & mAP \\
\Xhline{0.4pt}
Source only & 14.0 & 40.7 & 24.4 & 22.4 & 14.5 & 20.5 & 16.1 & 22.6 \\
\Xhline{0.4pt}
DA-Faster~\cite{DBLP:conf/cvpr/Chen0SDG18} & 14.3 & 44.6 & 26.5 & 29.4 & 15.8 & 20.6 & 16.8 & 24.0 \\
SW-Faster~\cite{DBLP:conf/cvpr/SaitoUHS19} & 15.2 & 45.7 & 29.5 & 30.2 & 17.1 & 21.2 & 18.4 & 25.3 \\
CR-DA-DET~\cite{DBLP:conf/cvpr/XuZJW20} & 19.5 & 46.3 & 31.3 & 31.4 & 17.3 & 23.8 & 18.9 & 26.9 \\
\Xhline{0.4pt}
SFOD-Mosaic~\cite{DBLP:conf/aaai/LiCXYYPZ21} & 20.6 & \textbf{50.4} & 32.6 & 32.4 & 18.9 & 25.0 & 23.4 & 29.0 \\
$\text{A}^2$SFOD(Ours) & \textbf{26.6} & 50.2 & \textbf{36.3} & \textbf{33.2} & \textbf{22.5} & \textbf{28.2} & \textbf{24.4} & \textbf{31.6} \\
\Xhline{0.4pt}
Oracle & 53.4 & 53.5 & 42.8 & 41.9 & 37.3 & 38.8 & 58.1 & 47.1 \\
\Xhline{0.4pt}
\end{tabular}
\end{center}
% \vspace{-0.3cm}
\caption{Adaptation to Large-Scale Dataset: Cityscapes $\rightarrow$ BDD100k}
\label{Tab:Cityscapes-BDD100k}
\end{table*}

\begin{figure*}[t]
\centering
\includegraphics[width=\linewidth]{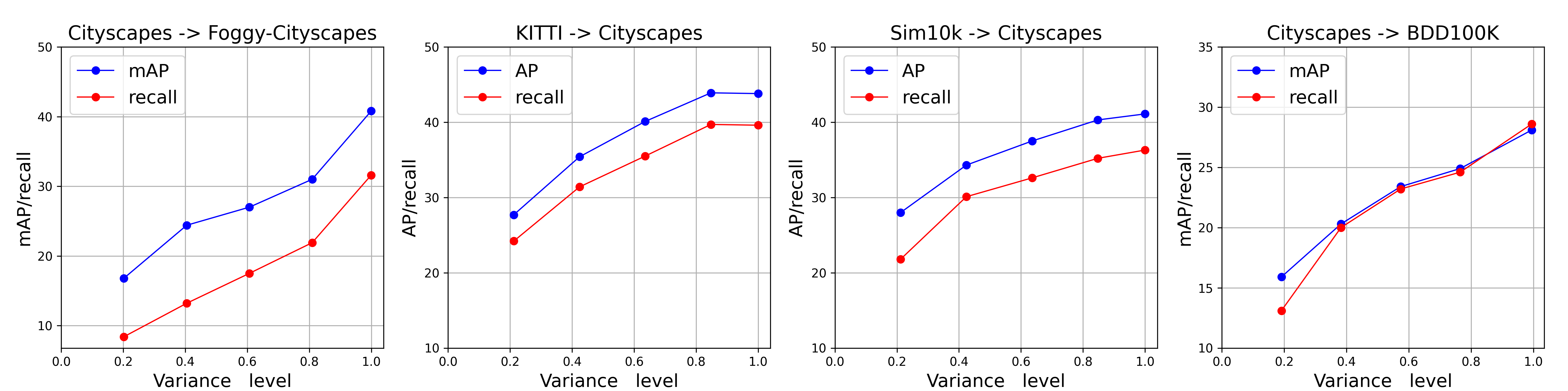}
\caption{mAP/recall-variance relation curves in four adaption tasks. We compute the variance of each image and split data into image groups by the level of their variance. We then measure the mAP/AP and the recall of these image groups. The recall is computed under the prediction confidence = 0.5.} 
\label{Fig:mAP curves}
% \vspace{-0.3cm}
\end{figure*}
\subsubsection{Adaptation from Normal to Foggy Weather}
Weather condition shift is very common in real-world applications, such as autonomous driving, which requires the strong transferability of models in different weathers, especially for the obscure objects caused by extreme weather. Thus, we employed Cityscapes as the source domain and Foggy-Cityscapes (a dataset in the foggy weather) as the target domain to benchmark the methods. The results of $\text{A}^2$SFOD and other methods are summarized in Table~\ref{Tab:Cityscapes-Foggy-Cityscapes}. Compared with the baseline method SFOD-Mosaic~\cite{DBLP:conf/aaai/LiCXYYPZ21}, we achieved a $1.9\%$ mAP score improvement on average. For a closer look at different classes, $\text{A}^2$SFOD obtained great success in the long-tail classes such as truck and train, while got limited improvement in the main classes such as car and bus. It is because our method which aligns the source-dissimilar images to source-similar ones encourages the model to give more confident detection. For the easy main classes, it may bring more false detection, while for long-tail classes, it effectively reduces the missing detection.

% \begin{figure}[h]
% \subfigure[Source images(not available)]
% {
%     \begin{minipage}[b]{.5\linewidth}
%         \centering
%         \includegraphics[scale=0.05]{images/source_1.jpg}
%     \end{minipage}
%     \begin{minipage}[b]{.5\linewidth}
%         \centering
%         \includegraphics[scale=0.05]{images/source_2.jpg}
%     \end{minipage}
% }
% \subfigure[Source similar images]
% {
%     \begin{minipage}[b]{.5\linewidth}
%         \centering
%         \includegraphics[scale=0.05]{images/source_similar_compare_1.jpg}
%     \end{minipage}
%     \begin{minipage}[b]{.5\linewidth}
%         \centering
%         \includegraphics[scale=0.05]{images/source_similar_compare_2.jpg}
%     \end{minipage}
% }
% \subfigure[Source dissimilar images]
% {
%  	\begin{minipage}[b]{.5\linewidth}
%         \centering
%         \includegraphics[scale=0.05]{images/source_dissimilar_compare_1.jpg}
%     \end{minipage}
%     \begin{minipage}[b]{.5\linewidth}
%         \centering
%         \includegraphics[scale=0.05]{images/source_dissimilar_2.png}
%     \end{minipage}
% }
% % \vspace{-0.2cm}
% \caption{Examples of source images, source similar images, and source dissimilar images in Foggy-Cityscapes. The source-similar images are more similar to the source images than the source-dissimilar ones.}
% \label{Fig:examples}
% % \vspace{-0.5cm}
% \end{figure}

\begin{figure}[h]
\subfigure[vl = 0.2]
{
    \centering
    \includegraphics[width=0.3\linewidth]{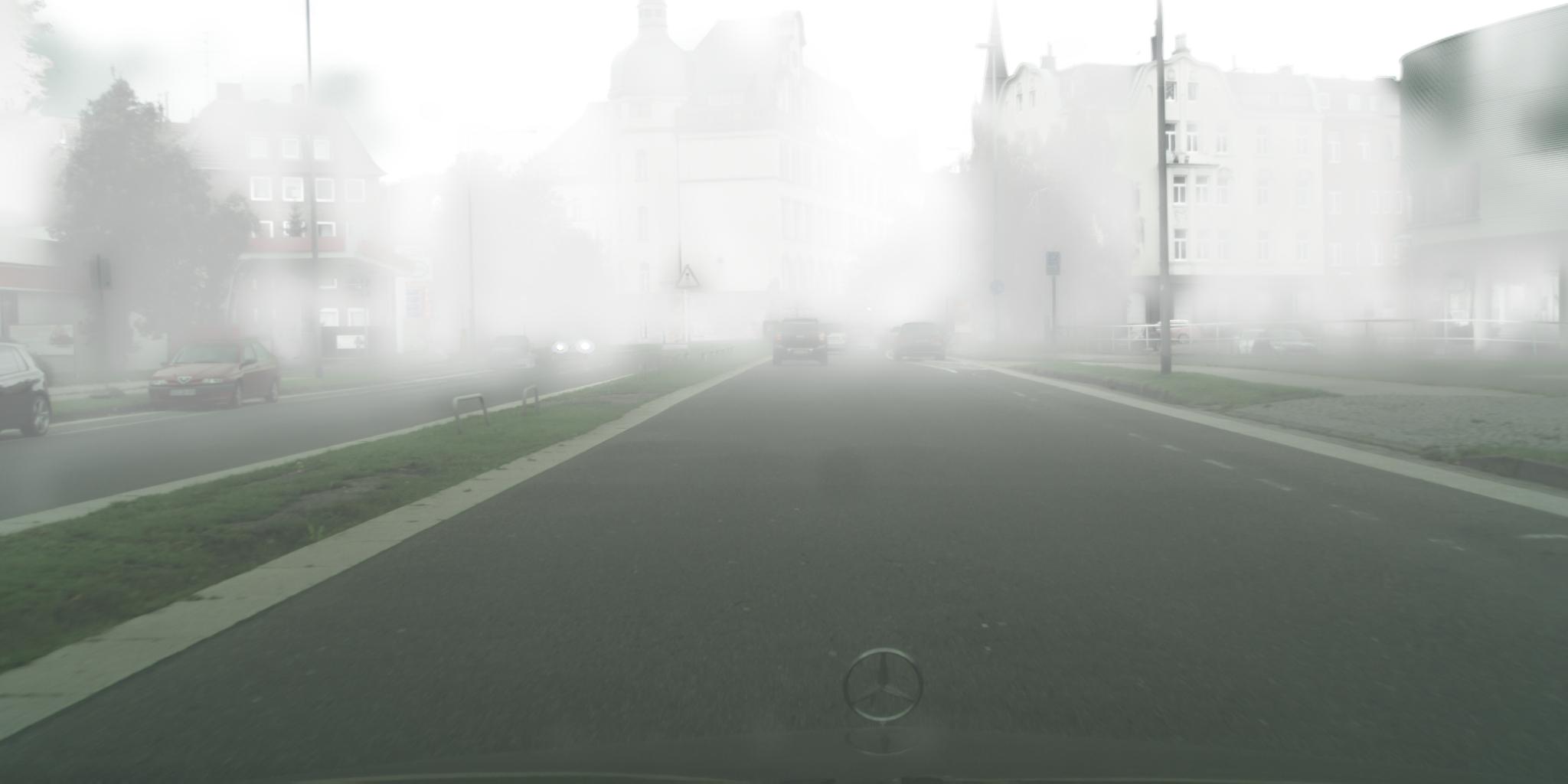}
    \label{subfig:4a}
}
\subfigure[vl = 0.4]
{
    \centering
    \includegraphics[width=0.3\linewidth]{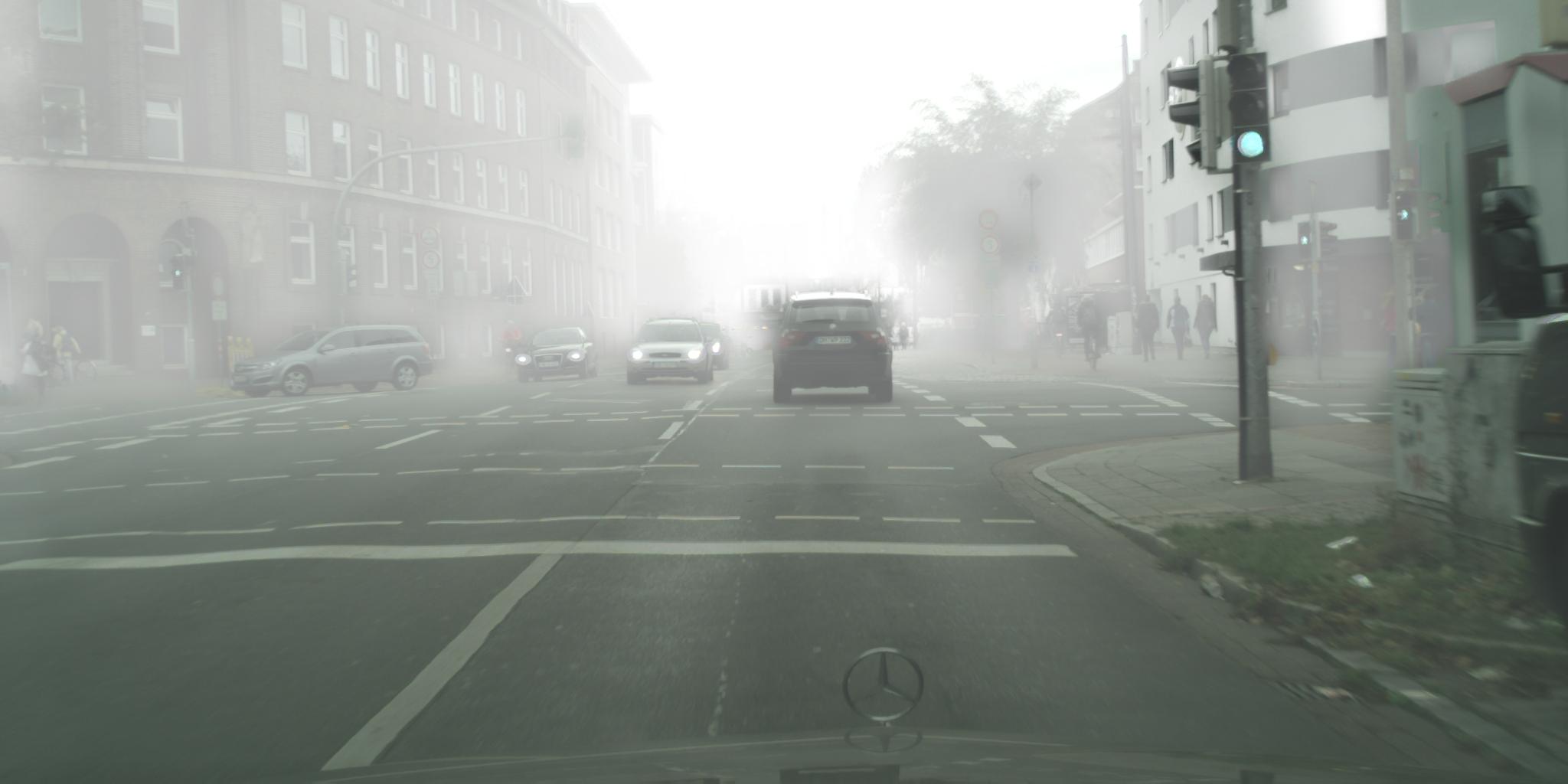}
    \label{subfig:4b}
}
\subfigure[vl = 0.6]
{
    \centering
    \includegraphics[width=0.3\linewidth]{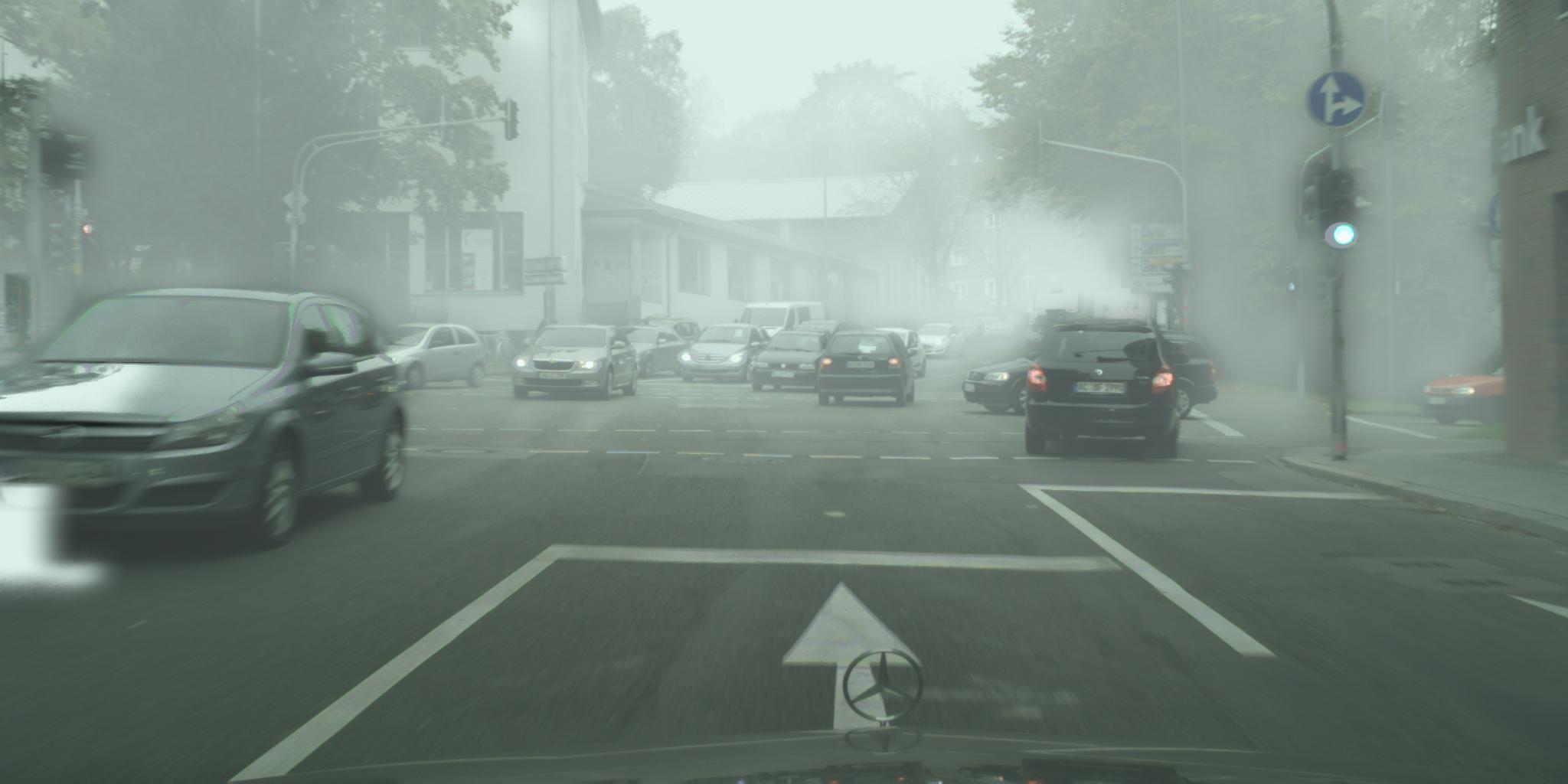}
    \label{subfig:4c}
}

\subfigure[vl = 0.8]
{
    \centering
    \includegraphics[width=0.3\linewidth]{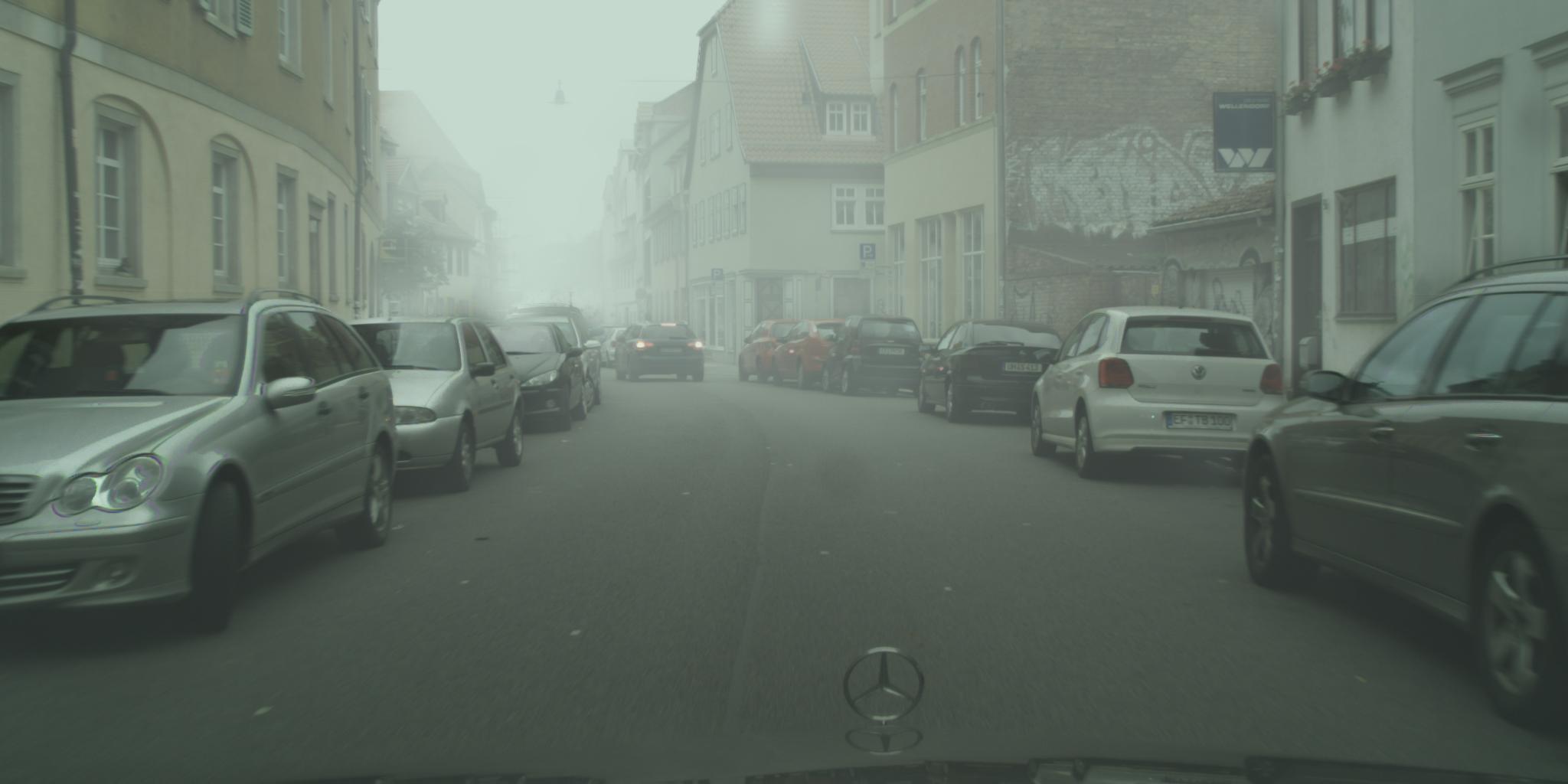}
    \label{subfig:4d}
}
\subfigure[vl = 1]
{
    \centering
    \includegraphics[width=0.3\linewidth]{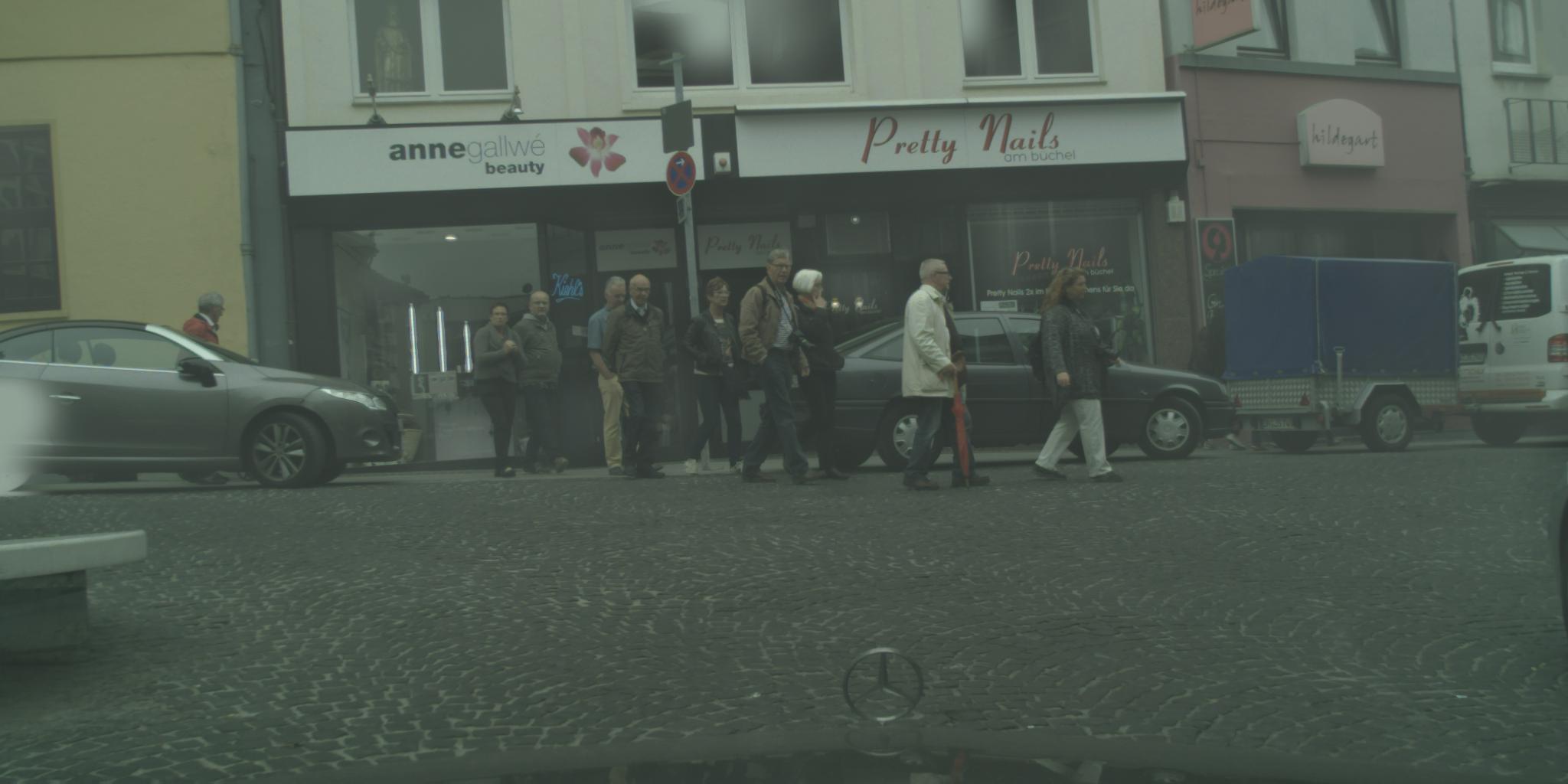}
    \label{subfig:4e}
}
\subfigure[source]
{
    \centering
    \includegraphics[width=0.3\linewidth]{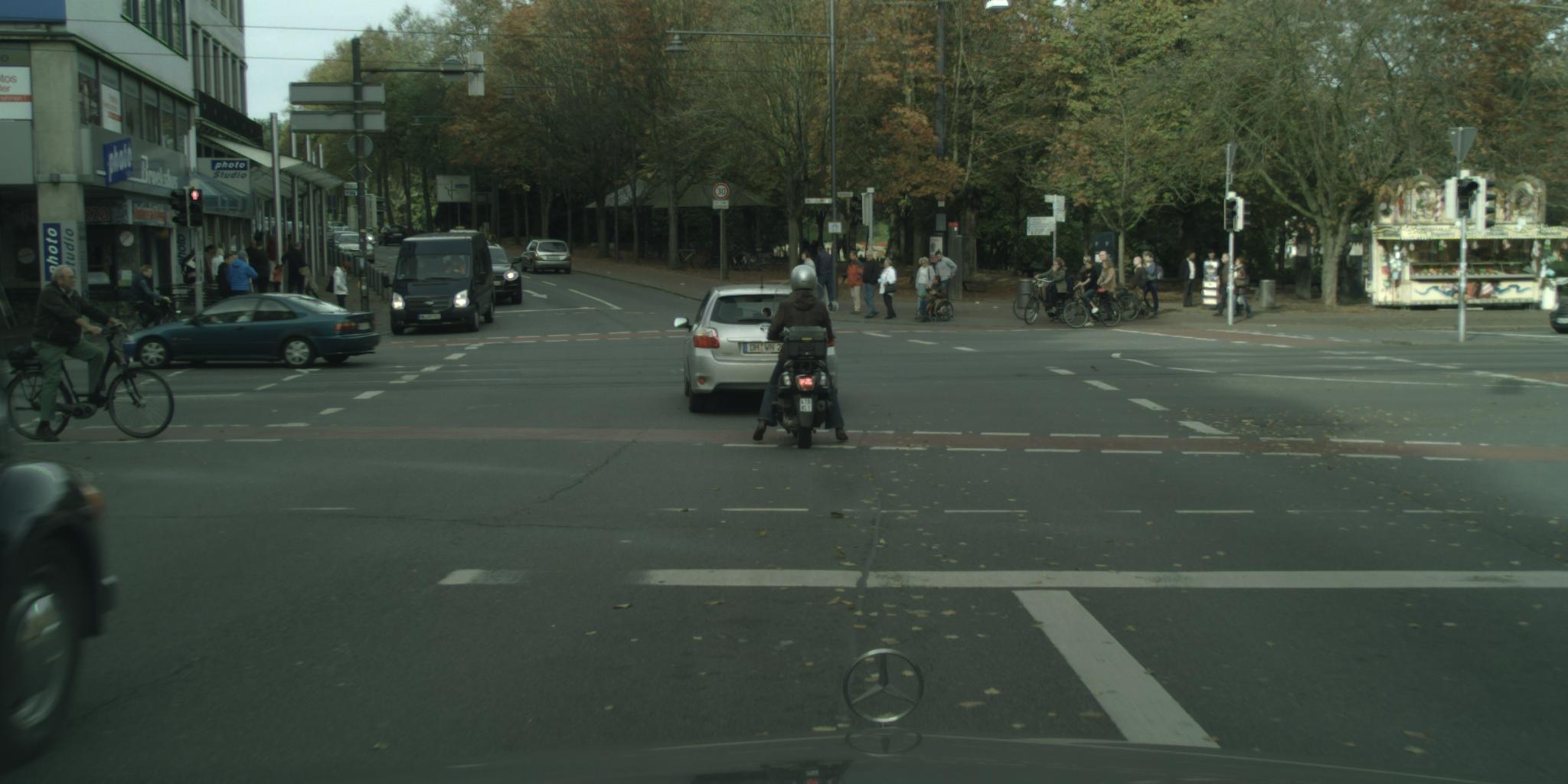}
    \label{subfig:4f}
}
\caption{Examples of images with different variance level from 0.2 to 1.0 and images of source domain.}
\label{Fig:examples}
\end{figure}

% change the order,  foggy setting first; it is the main experiments, the ablation study and the main story are built in this setting. 
\subsubsection{Adaptation to Unseen New Scenes} 
Besides, intelligent systems are always required to robustly adapt from a training environment to unseen new environments. To evaluate the transferability of our method to unseen new scenes, we measure the detection performance of methods that are trained on KITTI and tested on Cityscapes with camera setup differences. The experiment results are shown in Table~\ref{Tab:KITTI-Cityscapes}. We found that the performance improvement was not as significant as in other settings like weather changing. It may be because the variance of the target dataset, Cityscapes, is limited, and the difference between KITTI-similar and KITTI-dissimilar sets is not significant.

\subsubsection{Adaptation from Synthetic to Real Images} 
In autonomous driving, labeling real-world data is expensive and time-consuming. One potential solution is to simulate the real world and train the model with synthetic data. However, there is a large domain gap between the synthetic data and the real environment. It motivates us to transfer the knowledge learned from synthetic data to real images. In this experiment, we selected a synthetic dataset, Sim10k, as the source domain and Cityscapes as the target domain. As shown in Table~\ref{Tab:Sim10k-Cityscapes}, our method consistently outperforms the baseline SFOD-Mosaic~\cite{DBLP:conf/aaai/LiCXYYPZ21} and other methods.

\subsubsection{Adaptation to Large-Scale Dataset} 
Despite easily collecting large amounts of image data, the data annotation is expensive and labor intensive. Therefore, the transferability from limited labeled data to an unlabeled large-scale target dataset matters. In this experiment, we used Cityscapes as the source domain and BDD100k as the target domain to train the model and evaluated the detection results on only 7 common categories of the two datasets including ``truck'', ``car'', ``rider'', ``person'', ``motor'', ``bicycle'', and ``bus''. In Table~\ref{Tab:Cityscapes-BDD100k}, $\text{A}^2$SFOD obtained $31.6\%$ mAP, which achieves $+2.6\%$ mAP improvement than the existing best result.

\begin{table*}[tbp]
% \vspace{-0.3cm}
\begin{center}
% \small
\centering
\begin{tabular}{c|cccccccc|c}
\Xhline{0.4pt}
Methods & truck & car & rider & person & train & motor & bicycle & bus & mAP \\
\Xhline{0.4pt}
% Source only & 11.6 & 38.7 & 31.4 & 23.6 & 9.4 & 17.3 & 27.4 & 19.0 & 22.1 \\
% \Xhline{0.4pt}
Baseline & 20.0 & 43.7 & 37.8 & 26.6 & 13.0 & 24.8 & 37.1 & 36.4 & 29.9 \\
Baseline + MT & 23.7 & 44.0 & 42.9 & \textbf{32.5} & 12.9 & 29.7 & 38.1 & \textbf{37.0} & 32.6 \\
Baseline + MT + TSD (Our $\text{A}^2$SFOD) & \textbf{28.1} & \textbf{44.6} & \textbf{44.1} & 32.3 & \textbf{29.0} & \textbf{31.8} & \textbf{38.9} & 34.3 & \textbf{35.4} \\
\Xhline{0.4pt}
\end{tabular}
\end{center}
% \vspace{-0.5cm}
\caption{Ablation study with regarding to different components of $\text{A}^2$SFOD.}
\label{Tab:Ablation Study}
\end{table*}

\subsection{Ablation Studies and Further Analysis}

In this subsection, we conducted ablation studies to investigate the effectiveness of each component and gave more analysis. We are to answer the following questions.

\textbf{Q: Why the detection variance of the pretrained model can be regarded as a criterion for target division. A: It is because of a finding that the images with larger detection variances are more similar to the data in the source domain. 
} Target division aims to divide the target data into a source-similar set and a source-dissimilar set for aligning the target data and untouched source data. In this paper, we proposed to use detection variance as the criterion for target division since we found the larger detection variance denotes higher recall and more similar to the source data.
The pretrained model tends to give more predictions on the source-similar images. It causes a higher recall since more predictions mean less missed detection. On the other hand, more predictions indicate that some predictions are uncertain, which have a large variance with different dropout samplings.
We designed two experiments to verify our findings.

To verify the relation between the detection variance and the similarity to source data, we designed a qualitative experiment to show the source images and target images at different variance levels. While it is difficult to quantitatively identify this relation due to the lack of the similarity metric, we can clearly observe that the images with larger variance are more similar to the source data from Figure~\ref{Fig:examples}. As we set the threshold parameter $\sigma=0.7$, (a), (b) and (c) are considered to be source-dissimilar images and (d), (e) represent source-similar images.
% \textcolor{red}{The variance level represents the relative variance of an image compared with other target domain images. 

To verify whether the detection variance and recall are positively correlated, we provided a quantitative evaluation. Specifically, given a pre-trained model, we first sorted the testing data by the variance value and split them into several groups. Then we tested the model on the groups with different variance levels and calculated the recall score. As shown in Figure~\ref{Fig:mAP curves}, we conducted experiments on four settings including $Cityscapes\rightarrow Foggy-Cityscapes$, $KITTI\rightarrow Cityscapes-Car$, $Sim10k\rightarrow Cityscapes-Car$, and $Cityscapes\rightarrow BDD100k$, where $A\rightarrow B$ denotes that we pretrained the model on A and tested the model on B. In all settings, we found the variance level and recall score have highly positive correlations, with correlation coefficients $R^2 = 0.962,0.933,0.848,0.927$ respectively.

Besides, we also provided the mAP performance in the blue curves of Figure~\ref{Fig:mAP curves}, which shows that we can obtain better detection results in images with larger variances. Generally speaking, we can obtain better detection results with the source pretrained model on the source similar images, which also demonstrates the positive relation between large variance and source similarity. 

\textbf{Q: Can we directly apply the recall as the criterion?
A: No.} Recall is a metric that evaluates the detection of a set of images. However, the criterion is a metric for each image to divide the source-similar and source-dissimilar sets. When considering recall in a single image, the number of objects is variant and may mislead the division. 

\textbf{Q: What are the effects of each component in the method? 
A: Both target self-division (TSD) based alignment and Mean Teacher (MT) based fine-tuning are critical for our method.}
The main differences between $\text{A}^2$SFOD and other methods are the target self-division (TSD) based alignment and Mean Teacher (MT) based fine-tuning.
To explore the effectiveness of each component, we implemented an ablation study experiment on Cityscapes $\rightarrow$ Foggy-Cityscapes. The experiment results are summarized in Table~\ref{Tab:Ablation Study}. We remove the Mean-Teacher based fine-tuning and target self-division from $\text{A}^2$SFOD as our baseline. When adding the MT to the baseline, we can achieve
+2.7$\%$ mAP improvement. When applying TSD-based alignment, we can further achieve the 35.4$\%$ mAP score with +2.8$\%$ improvement.  
These significant improvements demonstrate that both components are critical for our method.

% \textbf{Q: Can Target Self-Division benefit from the bootstrap? A: No.}
% The source knowledge is stored in the pre-trained model, which helps to classify the source-similar and source dissimilar data. When we apply the bootstrap, i.e. update the model by target data, rich source information contained in the model will be suppressed. Such finetuned model cannot be applied for target self-division. Specifically, we can obtain the $29.6\%$ mAP with our original setting but only $29.4\%$ when further applying the bootstrap. Please see the supplementary material for more experimental results and analysis.

% \textbf{Q: Can we merge stage 3 and stage 4 by jointly finetuning and aligning? A: No.}
% Instead of our stage-by-stage training, we jointly trained adversarial alignment and mean teacher with both source-similar and source-dissimilar data. It leads to a performance drop from $33.5\%$ (stage-by-stage) to $31.5\%$ (joint). It may be because the high noises in pseudo labels of source-dissimilar data badly deteriorate the detection performance. It motivates us to first implement domain alignment to obtain high-quality pseudo labels over the whole target data and add a stage to finetune. More experimental results and analysis can be found in the supplementary material.

\section{Conclusion}
In this work, we have proposed an adversarial learning based method $\text{A}^2$SFOD to enable better object detection performance in source-free circumstances. The basic idea is to self-divide the target dataset into source-similar and source-dissimilar parts and align them in the feature space. The source-pretrained model can generate high-quality pseudo supervisions for the aligned target domain. To achieve this, we developed a detection variance-based self-division criterion. We evaluated our proposed method on five cross-domain object detection datasets. Experimental results show our superiority over compared SFOD methods as well as the effectiveness of each component. In the future, we will explore to integrate our proposed $\text{A}^2$SFOD to various detection backbones. 

\section{Acknowledgments}
This research was partly supported by the National Key R\&D Program of China (Grant No. 2020AAA0108303) \& NSFC 41876098, and by Shenzhen Science and Technology Project (Grant No. JCYJ20200109143041798) \& Shenzhen Stable Supporting Program (WDZC20200820200655001) \& Shenzhen Key Laboratory of next generation interactive media innovative technology (Grant No. ZDSYS20210623092001004).
\bigskip
\bibliography{aaai23}
\end{document}